# Advancing the Research and Development of Assured Artificial Intelligence and Machine Learning Capabilities

Tyler J. Shipp,[1] Daniel J. Clouse,[2] Michael J. De Lucia,[3] Metin B. Ahiskali,[4]
Kai Steverson,[5] Jonathan M. Mullin,[6] Nathaniel D. Bastian[7]

U.S. Army Combat Capabilities Development Command C5ISR Center,[1,4,5,6] National Security Agency Laboratory for Advanced Cybersecurity Research,[2] U.S. Army Combat Capabilities Development Command Army Research Laboratory,[3] U.S. Military Academy Army Cyber Institute,[7] U.S. Department of Defense Joint Artificial Intelligence Center[7]

tyler.j.shipp3.ctr@mail.mil,[1] djclous@evoforge.org,[2] michael.j.delucia2.civ@mail.mil,[3] metin.b.ahiskali.civ@mail.mil,[4] christopher.k.steverson.ctr@mail.mil,[5] jonathan.m.mullin.ctr@mail.mil,[6] nathaniel.bastian@westpoint.edu[7]

## Abstract

Artificial intelligence (AI) and machine learning (ML) have become increasingly vital in the development of novel defense and intelligence capabilities across all domains of warfare. An adversarial AI (A2I) and adversarial ML (AML) attack seeks to deceive and manipulate AI/ML models. It is imperative that AI/ML models can defend against these attacks. A2I/AML defenses will help provide the necessary assurance of these advanced capabilities that use AI/ML models. The A2I Working Group (A2IWG) seeks to advance the research and development of assured AI/ML capabilities via new A2I/AML defenses by fostering a collaborative environment across the U.S. Department of Defense and U.S. Intelligence Community. The A2IWG aims to identify specific challenges that it can help solve or address more directly, with initial focus on three topics: AI Trusted Robustness, AI System Security, and AI/ML Architecture Vulnerabilities.

## Introduction

Artificial intelligence (AI) and machine learning (ML) have become increasingly vital in the development of novel defense and intelligence capabilities across all domains of warfare. For U.S. Department of Defense (DoD) and Intelligence Community (IC) organizations, the security, robustness, resiliency and safety of AI and ML technologies is of great importance. Adversarial AI (A2I) and Adversarial ML (AML) methods deceive and manipulate AI and ML technologies. These adversarial methods have the potential to result in degraded performance of AI and ML capabilities, which can potentially cause catastrophic failures for missions that depend on them for success. Implementing assured AI and ML capabilities is paramount to ensuring they operate in a secure, robust, resilient, and safe manner for operational use and adoption on the battlefield.

In pursuit of advancing the research and development of assured AI and ML capabilities, the A2I Working Group (A2IWG) formed to provide a collaborative environment across DoD and IC organizations that have shared interests in advancing work in understanding and defending against A2I and AML. The A2IWG will help the DoD and IC to address challenges and solve them directly as appropriate.

There are many challenges along the path to assured AI and ML capabilities. These challenges span across basic research, applied research, development, fielding, and sustainment. Most of these challenges are ongoing research topics, while there is some advancement that has been made beyond just research and development. The A2IWG seeks to focus on issues that are of the greatest value to the community. Moreover, the A2IWG will continuously conduct activities and complete tasks that are well suited to collaborative problem solving, helping to avoid duplicating existing efforts and spreading awareness of relevant existing efforts.

The founding members of the A2IWG determined that AI Trusted Robustness, AI System Security, and AI/ML Architecture Vulnerabilities are the initial three topic areas for which the A2IWG will focus efforts for the DoD/IC. Specifically, work on AI Trusted Robustness seeks to explore and address challenges that exist in the overlap between Trustworthy AI and Adversarial Robustness. Work on AI System Security seeks to map out a holistic defense-in-depth approach to defending AI and ML capabilities from A2I and AML. Finally, work on AI/ML Architecture Vulnerabilities investigates the fundamentals of A2I and AML methods and how they interact with different AI and ML algorithms.

## AI Trusted Robustness

The concept of trust in computing can arguably be traced back to an origin with the National Academies of Science and their seminal publication Trust in Cyberspace in 1999 (NRC 1999), predating Bill Gates's famous "Trustworthy Computing" email he sent to every full-time employee at Microsoft in 2002 (Gates 2002). In that memo, he laid out four pillars: security, privacy, reliability and business integrity, where one generally acceptable definition of reliable is "performing at or exceeding expectations." Despite at least 30 years of existence as a concept, its goals can still be elusive and bugs are still found in modern software, regrettably sometimes catastrophically as evidenced by the Boeing 737 Max Lion Air disasters of 2019 (BBC 2019). Now, some say that we are in the middle of the "Summer of AI" and that it is due to the relatively recent advances in computer science hardware such as graphical processing units (GPUs). When many people refer to AI, they are actually referring to the ML subset of it, and its successes in computer vision have brought about a strong desire for DoD/IC organizations to leverage it. ML can be non-deterministic and this property is at least one reason why "Trustworthy AI" may be even more challenging to attain than Trustworthy Computing.

Many organizations across the DoD/IC enterprise are seeking transformational capabilities by leveraging AI/ML while striving to attain "Trustworthy AI" despite its challenges. The DoD, for example, made clear its desire to do so in its 2018 DoD AI Strategy: "We will invest in the research and development of AI systems that are resilient, robust, reliable, and secure; we will continue to fund research into techniques that produce more explainable AI; and we will pioneer approaches for AI test, evaluation, verification, and validation" (Mears, Clark and Bozman 2019). So, what is "Trustworthy AI" and how can DoD/IC organizations attain it? This is a hard question that could remain an open area of research for this century judging by "Trustworthy Computing's" 30+ years. Not surprisingly, many companies are eagerly claiming to provide solutions, for a fee, and one has apparently even gone as far as trade marking the name "Trustworthy AI" (Trustworthy AI, Inc. 2020). Nevertheless, the concept of Trustworthy AI is still in a state of flux; however, following Deloitte's lead, there are at least six pillars that are worthy of consideration for the subset that falls under ML (Deloitte 2020). Specifically, we should strive for ML implementations that are: 1) Robust/Reliable, 2) Safe/Secure, 3) Fair/Impartial, 4) Transparent/Explainable, 5) Protect Privacy and are 6) Responsible/Accountable.

Mitigating A2I/AML attack vulnerabilities is a very active area of research and development, is naturally an area of cybersecurity, and is also being invested in by DoD/IC organizations. A2I/AML research and development funded and spearheaded by organizations such as the Office of the Under Secretary of Defense for Research and Engineering (OUSD(R&E)), are charged with the development and oversight of DoD technology strategy for the DoD (OUSD 2020). Further, we can claim a success in advancing research and development in assured AI/ML capabilities via the Defense Advanced Research Projects Agency (DARPA) Guaranteeing AI Robustness Against Deception (GARD) program (Draper 2020), the DARPA Assured Autonomy program (Richards 2020), the DARPA Explainable AI (XAI) program (Turek 2020), as well as the through the Intelligence Advanced Research Projects Agency (IARPA) Trojans in AI (TrojAI) program (Alstott 2020). These organizations are actively striving to advance A2I/AML research and development, and it is these types of investments that will push the DoD/IC towards achieving the challenging and lofty goals of Trustworthy AI.

## AI System Security

Despite the end goal of secure AI/ML enabled capabilities, much of the research and development in A2I/AML defense focuses primarily on only the AI/ML model itself. Current research lacks focus on a defense for the entire AI/ML engineering pipeline. Common AI/ML engineering pipelines are composed of data collection, feature engineering, model training/validation/testing, and model deployment. Each stage of the AI/ML engineering pipeline must be protected against an adversarial attack in order to have a comprehensive defense. A2I/AML defenses are commonly focused on modalities applicable to computer vision and spam filtering. In contrast, there is a lesser focus on A2I/AML defenses on modalities applicable to cyber defense. Most A2I/AML defenses that perform well in computer vision and spam filtering applications have limited success in network defense applications (De Lucia and Cotton 2020; Alhajjar, Maxwell and Bastian 2020). Thus, many of the state-of-the-art defenses are not generically applicable to all AI/ML models.

A common goal of an A2I/AML attack is to cause misclassification by AI/ML models. In the cyber domain, for example, the attacker's A2I/AML objective is characterized by three common types of misclassifications. A Targeted False Negative misclassification attack misleads an AI/ML model to classify a malicious sample as benign (i.e., avoids detection) (De Lucia and Cotton 2020). The objective of a Targeted False Positive misclassification attack is to inhibit, deny or degrade the targeted AI/ML models by causing effects that deny the availability of valid responses from the models. Lastly, some targeted misclassifications are utilized by A2I/AML to cause specific reactions, within a capability or system that relies on the AI/ML model, that are desirable to the attacker.

Many investigations primarily focus on protections against evasion attacks, which focus on identifying inputs that produce misclassifications during the test phase (i.e., runtime data). Data poisoning attacks target earlier stages of AI/ML engineering pipeline by maliciously tampering with data collection, for example, which can have cascading effects on the following phases in the pipeline. These cascading effects can compromise the security of the resulting AI/ML model. Defenses that protect these earlier stages (data collection, feature engineering and training) from the effects of data poisoning are critical, as these types of attacks can degrade prediction quality or redirect predictions altogether (Devine and Bastian 2019). It is critical to consider the chain of custody and the source of data and its corresponding labels. In addition to evasion and poisoning attacks, attackers can infer information about training data, and attackers can approximately reconstruct the AI/ML model for further analysis and exploitation. A more comprehensive AML taxonomy is available from the National Institute of Standards and Technology (Tabassi et al. 2019).

Many AI/ML models integrate with larger systems, products or capabilities and require safeguards to protect the underlying AI/ML models from A2I/AML attacks. There is a vast amount of research in the cybersecurity domain to address system-level security concerns. Understanding and addressing these concerns and risks is critical to addressing A2I/AML defenses in the deployment phase. This is usually accomplished by processes such as the application of risk management frameworks, red teaming exercises, penetration tests and security audits. Similar processes that are focused on defending against and modeling A2I/AML attacks will need to be developed. Such an A2I/AML security evaluation of systems leveraging AI/ML will be essential in informing system engineers of A2I/AML vulnerabilities. Methods to conduct security evaluation of AI/ML models have not been sufficiently investigated to date. Risk identified by assessment vulnerabilities will need to have corresponding mitigations and defensive measures. This mapping of A2I/AML risks to defenses has not been sufficiently investigated either. Some suggest AI/ML designers should model and simulate an adversary, evaluate the impact, and develop countermeasures (Biggio and Roli 2018).

## AI/ML Architecture Vulnerabilities

It is important to explore the vulnerabilities that arise from different AI/ML model architectures, such as supervised learning, unsupervised learning, and reinforcement learning. An overview of known adversarial attacks highlights how different AI/ML model architectures lead to different types of vulnerabilities.

Attacks on supervised learning have been well documented via several avenues, especially in the computer vision domain. A general breakdown of attacks is through mimicry of representations with adversarial data (Feature Collision, Convex Polytype) or though minor perturbation which are then added to base images to collide in feature space (Clean Label Backdoor, Hidden Trigger Backdoor). These attacks have shown viability of the adversarial data examples, yet an understanding of how viability relates to real world risk requires quantification. Transferability of attacks is especially of interest for the government sector where models may not be openly available. Simple changes to the stochastic optimization can render previous examples non-viable (Schwarzschild et al. 2020). Statistical quantification of risk requires a uniform attack budget (% adversarial data). Stochastic optimization implies a need for statistical significance arguments rather than singular examples. Additionally, not all labeled classes are equally vulnerable, random sampling is required (Schwarzschild et al. 2020). Further, domains such as cyber, require a deeper understanding of when percent risk of adversarial attack is the best judge, versus a single catastrophic rare event attack.

As an unsupervised anomaly detection models determine a baseline of normal behavior and then looks for deviations from that baseline; for example, identifying malicious activity on a computer network by looking for unusual behavior. Attacks against anomaly detectors typically target the AI/ML model's understanding of what is normal. Poisoning attacks seek to expand the definition of normal to include malicious activity, while evasion attacks seek to modify malicious activity to fit the definition of normal. Anomaly detectors rely on a low amount of malicious activity occurring at training time. If a computer is already full of viruses, then they will be incorporated as part of the baseline normal. Anomaly detection models that are continuously trained on live data are susceptible to "boiling frog" attacks where the training data is gradually altered to be more and more anomalous (Rubinstein et al. 2009). This poisons the model while avoiding setting off alerts by presenting data that the current models finds anomalous.

Reinforcement learning can be used in environments where the agent can observe the state of the environment, take an action, and then receive feedback. For example, reinforcement learning is commonly used to learn how to play a video game. The agent observes the state of the game, then chooses an action from legal moves of the game. Feedback is typically provided based on the score or who won. Attacks against reinforcement learning models have been developed in two forms. First, there are attacks that manipulate the agent's observations (Lin et al. 2017). For example, altering the display information from a video game can trick the agent into the incorrect action. The second avenue of attack is to tailor an adversarial opponent use blind spots or gaps in the agent's training. For example, in a virtual racing program an adversarial opponent discovered that simply falling down in unusual ways caused the other agent to malfunction and not finish the race (Gleave et al. 2019).

## Conclusions

While many areas in A2I/AML research and development still need continued advancement, the current DARPA and IARPA programs have been leading the way across the DoD/IC enterprise in advancing the state-of-the-art. In addition to these two organizations, the OUSD(R&E) has several science and technology modernization research and development efforts underway in the domains of AI/ML and autonomy making advances in critical areas to include AI/ML verification and validation, AI robustness and resiliency, ethical AI, and other areas related to A2I/AML.

Given these recent advancements, the A2IWG has many different topics to deep dive. The topics of AI Trusted Robustness as a subset of Trustworthy AI, AI System Security, and AI/ML Architecture Vulnerabilities will continue to be scoped and refined so that more specific tasks can be identified that the A2IWG can address collaboratively. Through A2IWG collaborative meetings, DoD/IC organizations will be able to meet with each other and identify transition opportunities and partnerships between efforts. While different efforts may have different goals or use cases, often they both rely on some of the same underlying foundational concepts.

## Acknowledgments

The views expressed in this paper are those of the authors and do not reflect the official policy or position of the United States Army, the United States Department of Defense, or the United States Government.

## References

Alhajjar, E.; Maxwell, P.; and Bastian, N. 2020. Adversarial Machine Learning in Network Intrusion Detection Systems. arXiv preprint. arXiv: 2004.11898v1 [cs.CR]. Ithaca, NY: Cornell University Library.

Alstott, J. 2020. Trojans in Artificial Intelligence (TrojAI) Program, Intelligence Advanced Research Projects Agency (IARPA). Available from https://www.iarpa.gov/index.php/research-programs/trojai.

BBC. 2019. Boeing 737 Max Lion Air Crash Caused by Series of Failures. Available from https://www.bbc.com/news/business-50177788.

Biggio, B. and Roli, F. 2018. Wild patterns: Ten years after the rise of adversarial machine learning. *Pattern Recognition*, 84: 317–331. https://doi.org/10.1016/j.patcog.2018.07.023.

De Lucia, M. and Cotton, C. 2020. A network security classifier defense: against adversarial machine learning attacks. In Proceedings of the 2nd ACM Workshop on Wireless Security and Machine Learning. New York: Association for Computing Machinery. https://doi.org/10.1145/3395352.3402627.

Deloitte. 2020. Deloitte AI Institute: Bridging the Ethics Gap Surrounding AI. Available from https://www2.deloitte.com/us/en/pages/deloitte-analytics/solutions/ethics-of-ai-framework.html.

Devine, S. and Bastian, N. 2019. Intelligent Systems Design for Malware Classification Under Adversarial Conditions. arXiv preprint. arXiv: 1907.03149v1 [cs.LG]. Ithaca, NY: Cornell University Library.

Draper, B. 2020. Guaranteeing AI Robustness Against Deception (GARD) Program, Defense Advanced Research Projects Agency (DARPA). Available from https://www.darpa.mil/program/guaranteeing-ai-robustness-against-deception.

Gates, B. 2002. Trustworthy Computing. Available from https://www.wired.com/2002/01/bill-gates-trustworthy-computing/.

Gleave, A.; Dennis, M.; Wild, C.; Kant, N.; Levine, S.; and Russell, S. 2019. Adversarial Policies: Attacking Deep Reinforcement Learning. arXiv preprint. arXiv:1905.10615v2 [cs.LG]. Ithaca, NY: Cornell University Library.

Lin, Y.; Hong, Z.; Liao, Y.; Shih, M.; Liu, M.; and Sun, M. 2017. Tactics of Adversarial Attack on Deep Reinforcement Learning Agents. arXiv preprint. arXiv:1703.06748v4 [cs.LG]. Ithaca, NY: Cornell University Library.

Mears, Z.; Clark, S.; and Bozman, J. 2019. Defense Department Releases Artificial Intelligence Strategy. Available from https://www.insidegovernmentcontracts.com/2019/02/defense-department-releases-artificial-intelligence-strategy/.

National Research Council (NRC). 1999. Trust in Cyberspace. Washington, DC: The National Academies Press. https://doi.org/10.17226/6161.

OUSD. 2020. Office of the Under Secretary of Defense, Research and Engineering. Available from https://www.cto.mil/.

Richards, R. 2020. Assured Autonomy (AA) Program, Defense Advanced Research Projects Agency (DARPA). Available from https://www.darpa.mil/program/assured-autonomy.

Rubinstein, B.; Nelson, B.; Huang, L.; Joseph, A.; Lau, S.; Rao, S.; Taft, N.; and Tygar, J. 2009. ANTIDOTE: Understanding and Defending against Poisoning of Anomaly Detectors. In Proceedings of the 9th ACM SIGCOMM Conference on Internet Measurement. New York: Association for Computing Machinery. https://doi.org/10.1145/1644893.1644895.

Schwarzschild, A.; Goldblum, M.; Gupta, A.; Dickerson, J.; and Goldstein, T. 2020. Just How Toxic Is Data Poisoning? A Unified Benchmark for Backdoor and Data Poisoning Attacks. arXiv preprint. arXiv:2006.12557v1 [cs.LG]. Ithaca, NY: Cornell University Library.

Tabassi, E.; Burns, K.; Hadjimichael, M.; Molina-Markham, A.; and Sexton, J. 2019. A Taxonomy and Terminology of Adversarial Machine Learning, NISTIR 8269 (Draft). Gaithersburg, MD: National Institute of Standards and Technology. doi:10.6028/NIST.IR.8269-draft.

Trustworthy AI, Inc. 2020. Quantifying Risk with Unparalleled Efficiency. Available from https://trustworthy.ai.

Turek, M. 2020. Explainable AI (XAI) Program, Defense Advanced Research Projects Agency (DARPA). Available from https://www.darpa.mil/program/explainable-artificial-intelligence.